\documentclass[a4paper,fleqn]{cas-dc}

\usepackage{bbding}
\usepackage{underscore}
\usepackage[numbers]{natbib}
\usepackage{orcidlink}
\usepackage{subfig}
\def\tsc#1{\csdef{#1}{\textsc{\lowercase{#1}}\xspace}}
\tsc{WGM}
\tsc{QE}

\begin{document}
\let\WriteBookmarks\relax
\def\floatpagepagefraction{1}
\def\textpagefraction{.001}

\shorttitle{Shape-Prior-Based Point Cloud Completion for Single-Stage Fully Sparse 3D Object Detection}    

\shortauthors{Kaizheng Wang et al.}  

\title [mode = title]{Shape-Prior-Based Point Cloud Completion for Single-Stage Fully Sparse 3D Object Detection}  

\author[1]{Kaizheng Wang}[orcid=0009-0004-1732-6688,style=chinese]
\cormark[1]
\ead{wangkaizheng@njust.edu.cn}

\author[1]{Mingqian Ji }[style=chinese]
\ead{mingqianji@njust.edu.cn}

\author[1]{Jian Yang}[style=chinese]
\ead{csjyang@njust.edu.cn}

\author[1]{Shanshan Zhang}[style=chinese]
\cormark[1]
\ead{shanshan.zhang@njust.edu.cn}

\cortext[cor1]{Corresponding authors}

\affiliation[1]{organization={School of Computer Science and Engineering, Nanjing University of Science and Technology},
            city={Nanjing},
            country={China}}

\begin{abstract}
Single-stage fully sparse 3D object detectors rely on point clouds data to detect objects in autonomous driving scenarios. However, the sparsity and incompleteness of point clouds significantly limit the performance of 3D object detection. To address this issue, this paper proposes a point clouds completion method specifically designed for single-stage fully sparse detectors. The entire shape-prior-based completion process consists of two consecutive steps.
In the first step, we design a novel Instance Selection module, which is capable of identifying point clouds corresponding to foreground objects even when the baseline model does not generate proposals, while effectively ignoring the point clouds of background regions. In the second step, we introduce a novel Alignment-Based Point Completion module, which aligns the point clouds of foreground objects with prototypes in terms of both their centers and orientations. Subsequently, points are selected from the prototype to fill in the missing parts of the foreground object.
We evaluated our method on two single-stage fully sparse detectors using the KITTI dataset. The experimental results demonstrate that the proposed method significantly improves the detection performance, confirming its effectiveness and generalizability.
\end{abstract}


\maketitle

\section{Introduction}
\label{sec1}

LiDAR-based 3D object detection plays a critical role in autonomous driving systems. In recent years, numerous studies have focused on improving the performance of 3D detection. Recently, several single-stage fully sparse 3D object detection methods \cite{FSD,FSDV2,VoxelNext,SSD,SA_SSD,10438399,Yan_2024} have been proposed, which extract features by designing diverse network architectures and adopting different data representations (e.g., voxels and keypoints). These methods have demonstrated excellent detection performance, but have reached a bottleneck. The fundamental issue lies in the fact that existing research largely overlooks the impact of point clouds sparsity and incompleteness on detection performance.

To address this problem, some studies \cite{BTC,S2D,SSN,OKGR,SIE} have conducted in-depth analyzes of the challenges posed by the sparsity and incompleteness of raw point clouds data and have attempted to optimize two-stage detectors through point cloud or voxel completion. For example, SSN \cite{SSN} and SIENet \cite{SIE} exploit the self-correlation properties of objects and employ self-supervised learning to recover missing point cloud information, thereby enhancing the detection network’s ability to recognize and localize objects. However, these methods require dense representations to be completed within the second-stage network, which is computationally expensive and time-consuming. BtcDet \cite{BTC} and Sparse2Dense \cite{S2D}, on the other hand, predict shape occupancy information to complete the voxel in occluded regions, while OKGR \cite{OKGR} predicts key point offsets to generate points that approximate actual points in occluded areas. Although these approaches focus on sparse representations, they still rely on two-stage detectors, which compromises computational efficiency. Furthermore, the aforementioned methods are primarily designed for two-stage 3D object detectors and cannot be directly applied to single-stage fully sparse 3D object detectors.

To address the above issues, this paper proposes a shape-prior-based point cloud-level completion method, which is compatible with both voxel-based and point-based fully sparse detectors. Specifically, our approach consists of two core modules: Instance Selection and Alignment-Based Point Completion. In the Instance Selection stage, we design a point cloud classification module to distinguish points of different categories and propose a clustering module to group points within each category.In the Alignment-Based Point Completion stage, an alignment module is introduced to align the generated prototypes with the clustered instances. This is followed by a point completion module to complete the point clouds for each instance. 

\begin{figure*}[pos=htb]
    \centering
    \subfloat[Feature visualization of baseline]{\includegraphics[width=0.45\linewidth, trim=0 0 0 0, clip]{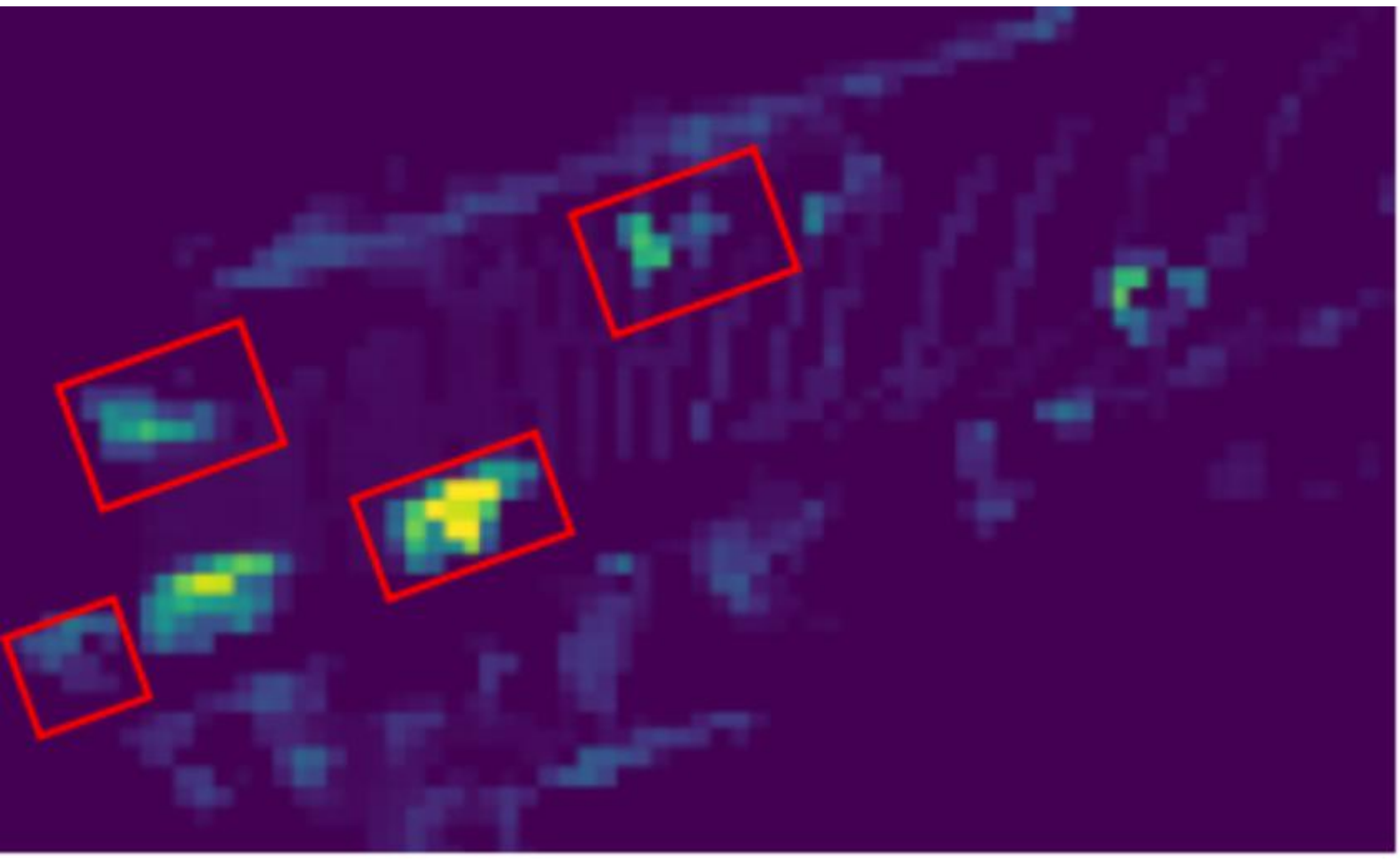}\label{fig_1_case}}
    \hspace{0.05\linewidth} 
    \subfloat[Feature visualization of ours]{\includegraphics[width=0.451\linewidth, trim=0 0 0 0, clip]{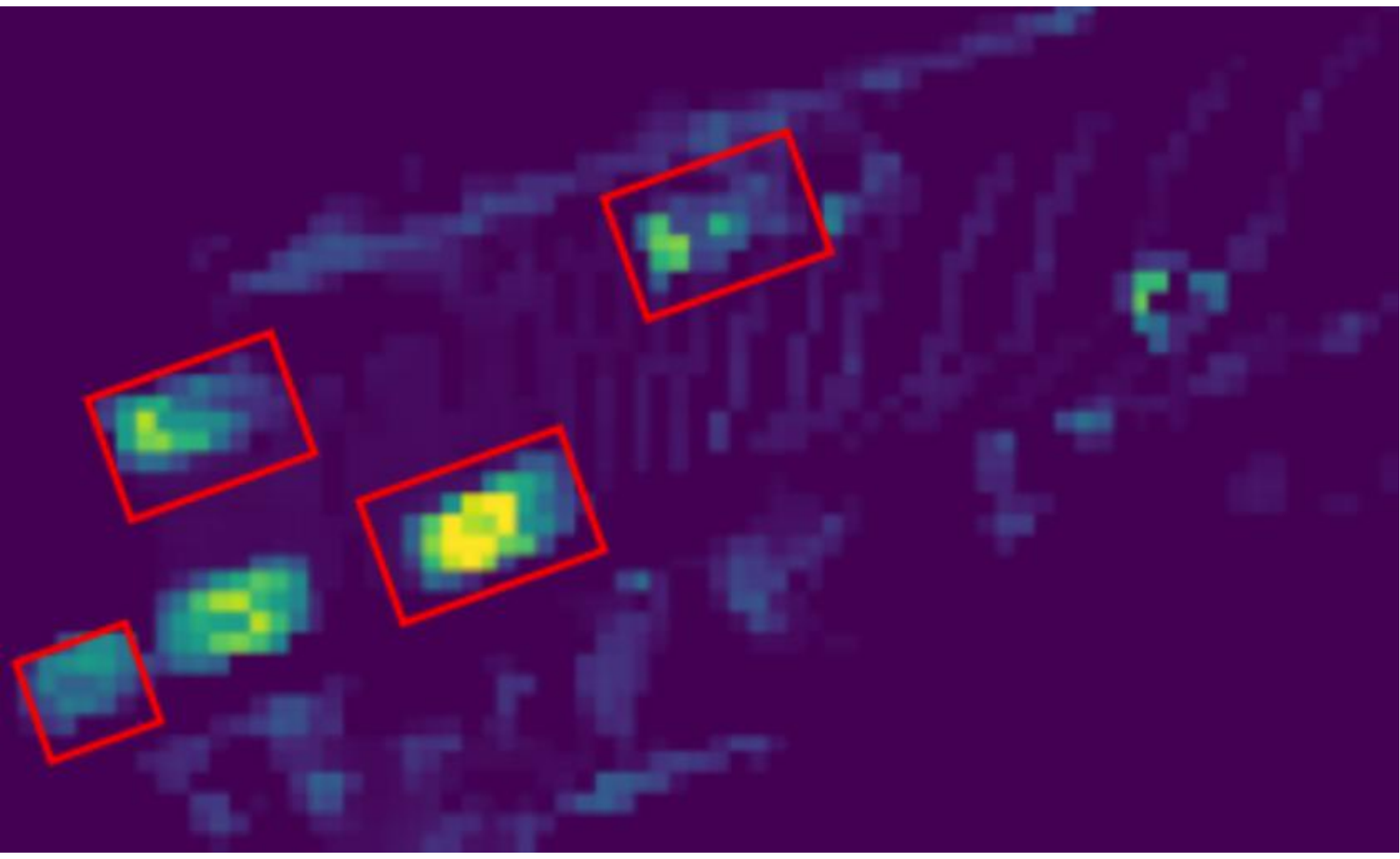}\label{fig_2_case}}
    \caption{The feature visualizations of the baseline method and our method. (a) and (b) respectively show the feature visualizations of the baseline method and our method in BEV. The lighter the color, the higher the feature quality. The \textcolor{red}{red boxed} areas highlight regions with significant differences.}
    \label{fig_1}
\end{figure*}

We apply the proposed method to state-of-the-art fully sparse detectors and achieve significant performance improvements. Fig. \ref{fig_1} compares the feature distributions between the baseline methods and ours, clearly demonstrating that our approach enhances the features in occluded regions, thereby improving the detection of heavily occluded objects.

The main contributions of this paper are summarized as follows:
\begin{itemize}
    \item We propose a point cloud completion method tailored for single-stage fully sparse detectors.
    \item We use an Instance Selection module to obtain instance point clouds and an Alignment-Based Point Completion module to perform point cloud completion, both of which are independent of proposal boxes.
    \item The proposed method is applied to two state-of-the-art fully sparse 3D object detectors: FSDV2 \cite{FSDV2} and VoxelNeXt \cite{VoxelNext}, and evaluated on the KITTI \cite{KITTI} validation and test sets. The experimental results demonstrate significant performance improvements.
\end{itemize}

\section{Related Works}
Since this study employs LiDAR-based fully sparse detectors as the foundational detection framework and further designs a completion module on this basis, we review the research progress in two related areas: LiDAR-based 3D object detection and point cloud completion.
\subsection{LiDAR-based 3D Object Detection}
In recent years, LiDAR-based 3D object detection techniques have made remarkable progress. Early two-stage methods \cite{PVRCNN,PVRCNN++,centerpoint,Voxelset,point2seq,PDV} achieved outstanding performance in 3D object detection tasks. These methods typically generate Bird's-Eye View (BEV) features of the scene using dense feature extraction networks, followed by two main strategies for detection: leveraging a Region Proposal Network (RPN) to produce candidate bounding boxes \cite{PVRCNN,PVRCNN++,centerpoint,Voxelset,PDV}, and encoding and fusing features further to predict the final 3D bounding boxes \cite{centerpoint,point2seq}.

However, the main drawback of these two-stage approaches lies in their high computational complexity and slow inference speed, making them unsuitable for real-time applications. To address these limitations, recent studies \cite{FSD,FSDV2,VoxelNext,SSD,SA_SSD} have proposed single-stage fully sparse detection methods based on fully sparse feature extraction networks. These methods predict 3D bounding boxes directly on sparse 3D feature maps, significantly improving detection speed while maintaining accuracy comparable to that of two-stage approaches. This advancement has established fully sparse detectors as the mainstream solution for 3D object detection tasks.However, these methods fail to account for the impact of point cloud sparsity and incompleteness on detector performance, whereas our approach explicitly addresses this issue.

\subsection{Point Cloud Completion}

Point cloud completion aims to refine sparse and incomplete point cloud data to improve the performance of 3D object detection. Existing methods can be categorized into two main types: dense representation completion methods \cite{SSN,SIE,proxyformer} and sparse representation completion methods \cite{SPG,BTC,S2D,OKGR}. 

Dense representation completion methods focus primarily on completing dense raw point clouds. For example, SSN \cite{SSN} introduces a shape feature network to complete raw point clouds, SIENet \cite{SIE} utilizes pre-trained models for point cloud completion, and ProxyFormer \cite{proxyformer} proposes an innovative positional encoding method that converts truly missing regions into proxies and predicts these proxies to achieve point cloud completion. These methods significantly improve detection performance, but the completion process for large-scale scenes still incurs high computational costs and is time-consuming. Moreover, these methods cannot be directly applied to single-stage fully sparse detectors.

Sparse representation completion methods focus on completing sparse point clouds or representations at the feature level. For instance, SPG \cite{SPG} employs a supervised learning strategy to predict shape occupancy and thereby achieve voxel-level completion; BtcDet \cite{BTC} introduces an auxiliary task in the detection framework to predict shape occupancy and complete voxel data; Sparse2Dense \cite{S2D} utilizes knowledge distillation techniques to enhance voxel-level completeness; OKGR \cite{OKGR}, based on the principle of central symmetry, generates points within candidate boxes and further predicts offsets to align the generated points closely with the real points in occluded areas. Although these methods effectively achieve sparse representation completion by accurately predicting shape occupancy, they still rely on two-stage networks, are computationally expensive, and depend on proposal boxes, making them difficult to integrate with single-stage fully sparse detectors. In contrast, our method completes sparse representations directly within a single-stage fully sparse network.
\section{Methodology}
\begin{figure*}[pos=htb]
    \centering
    \includegraphics[width=\linewidth]{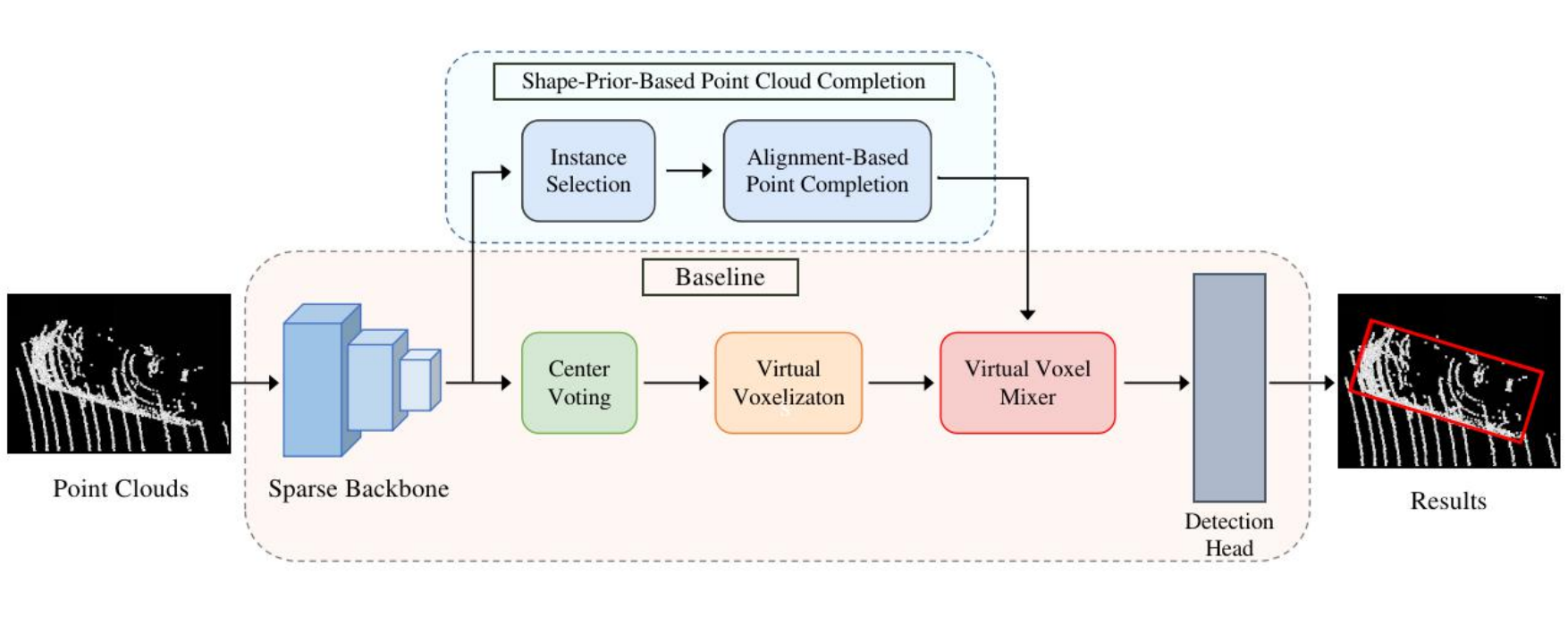}
    \caption{The overview of Shape-Prior-Based Point Cloud Completion for Single-Stage Fully Sparse 3D Object Detection (SPPCC). Our approach mainly consists of two modules, namely Instance Selection and Alignment-Based Point Completion.In the BEV view,the \textcolor{red}{red boxed} highlight the predicted bounding boxes.}
    \label{fig_2}
\end{figure*} 
Based on the aforementioned motivations and related work, we propose a completion method tailored for fully sparse 3D object detectors to enhance detection performance.
\subsection{Overview}
As illustrated in Fig. \ref{fig_2}, our method is applied to fully sparse detectors and consists of two primary modules: Instance Selection and Alignment-Based Point Completion. The overall workflow can be summarized as follows: first, the point cloud is passed through a sparse backbone network \cite{sparseunet} to extract features for each point. These features are then fed into the Instance Selection module to generate clustered instances. Next, each clustered instance is processed by the Alignment-Based Point Completion module to produce a completed point cloud. Details of the Instance Selection module are described in Sec. \ref{Instance Selection}, while the Alignment-Based Point Completion module is detailed in Sec. \ref{Alignment-Based Point Completion}.
 
 Additionally, the point features are input into the center voting network, where each point predicts its perceived object center. These predicted
center points are mixed with the completed point cloud features at the feature level. Finally, the resulting features are fed into the detection head to obtain 3D bounding boxes and corresponding confidence scores.

\subsection{Instance Selection}
\label{Instance Selection}
Inspired by recent point cloud completion methods \cite{OKGR,BTC,SSN} in two-stage detectors, the typical approach is to generate proposal boxes to locate objects and subsequently complete the points within these proposals. However, single-stage fully sparse methods and do not produce proposal boxes, making previous approaches inapplicable. To address this limitation, we adopt Instance Selection to identify objects in the scene.

\begin{figure}[pos=htb]
\centering
\includegraphics[width=1\linewidth]{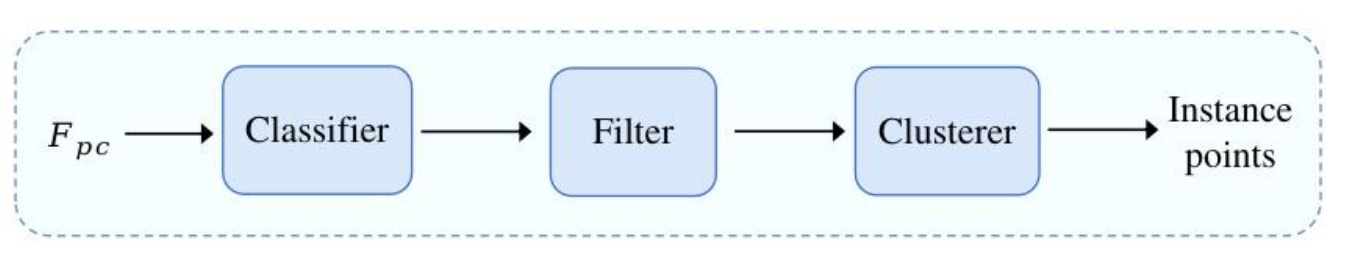}
\caption{Instance Selection module. It consists of three components: a classifier, a filter, and a clusterer.}
\label{fig_3}
\end{figure}

As shown in Fig. \ref{fig_3}, the Instance Selection module comprises a classifier, a filter, and a clusterer. Inspired by the recent fully sparse 3D object detection algorithm FSDV2 \cite{FSDV2}, we utilize a MLP layer as a classifier to predict the category scores for each point.

The classifier takes the feature set of points $ F_{pc} $ as input and outputs the corresponding class scores. After obtaining the class scores for all points, the filter removes background points based on a predefined threshold, leaving the class scores of foreground points. Subsequently, the points are grouped into different categories, and the classified points are fed into the clusterer to generate clustering instances (i.e., objects in the scene).
\subsection{Alignment-Based Point Completion}
\label{Alignment-Based Point Completion}
The clustered instances obtained in the previous section do not fully capture the complete shapes of the target objects. Therefore, we design a point completion scheme. Before proceeding, it is essential to prepare a prototype point cloud that represents the complete shape of the target object.

As shown in Fig. 4, the architecture of the Alignment-Based Point Completion module comprises two main steps: first, aligning the clustered instances with the prototype point cloud; second, filling the missing regions of the clustered instances with points selected from the prototype based on a completion coefficient ($CC$).

\textbf{Alignment}. The alignment process consists of two steps. In the first step, the network predicts the offset of each instance's centroid to determine the instance center, and the instance is then translated to align its center with the prototype's center. In the second step, the network predicts the yaw angle of the instance, and the instance is rotated to achieve full alignment with the prototype.

\begin{figure*}[!h]
    \centering
    \includegraphics[width=1\linewidth]{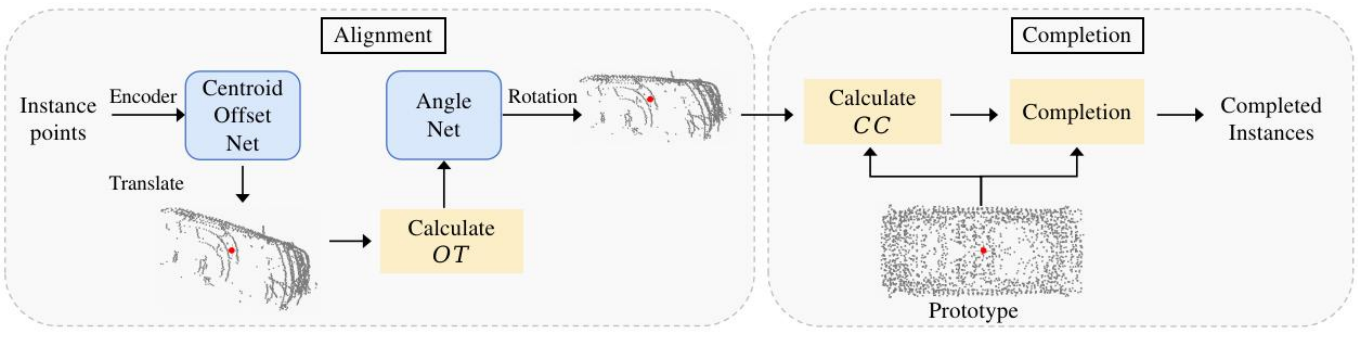}
    \caption{Alignment-Based Point Completion. It consists of two parts: Alignment and Completion. Alignment includes CenterNet and AngleNet.}
    \label{fig_4}
\end{figure*}

\begin{figure}[pos=htb]
    \centering
    \subfloat[]{\includegraphics[width=0.4\linewidth, trim=0 0 0 0, clip]{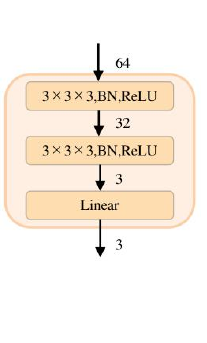}\label{fig_3_case}}
    \hspace{0.05\linewidth} 
    \subfloat[]{\includegraphics[width=0.4\linewidth, trim=0 0 0 0, clip]{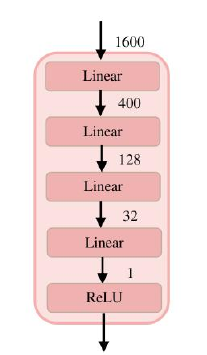}\label{fig_4_case}}
    \caption{The architecture of networks in Alignment.(a): Centroid Offset Network. (b): Angle Net.}
    \label{fig_5}
\end{figure}

Specifically, as illustrated in Fig. \ref{fig_4}, the instance point cloud is first encoded into instance features using a voxel encoder, which are then fed into the Centroid Offset Network. The Centroid Offset Network consists of two SparseConv blocks and a linear layer, where each SparseConv block includes a SparseConv3D, BatchNorm1D, and ReLU operation, as illustrated in Fig. \ref{fig_5}. The Centroid Offset Network predicts the centroid offset. Subsequently, the instance point cloud is translated to the origin to align its center with the prototype center based on the predicted offset.

Subsequently, a 2D Occupancy Table ($OT$) is computed from the point cloud, where a 2D occupancy map is generated based on the \(x\) and \(y\) coordinates of the point cloud. This occupancy map encodes the yaw angle information of the instance. The generated occupancy map is then fed into the Angle Network, which consists of four linear layers and one ReLU layer, to predict the yaw angle, as illustrated in Fig. \ref{fig_5}. Finally, the instance point cloud is rotated to align with the prototype point cloud.

\textbf{Completion}. Inspired by prior point cloud completion methods \cite{OKGR,BTC,SSN} in two-stage detectors, points have previously been generated symmetrically around the centroid or axes of the proposal boxes. These generated points are constrained within the proposal box and participate in subsequent prediction of the box's category and score. However, in fully sparse methods, no proposal boxes are generated, making it impossible to directly apply symmetry-based completion.

Upon analysis, we observe that the essence of symmetry-based completion lies in the relationship between the number of original and generated points: the more original points, the more generated points; the fewer original points, the fewer generated points.  Therefore, we design a completion coefficient ($CC$), calculated as follows:

\begin{equation}
CC = \frac{N_I}{N_p},
\end{equation}
where $N_I$ represents the number of points in the instance point cloud, while $N_p$ denotes the number of points in the prototype point cloud.

As illustrated in Fig. 4. The completion process is as follows: the aligned instance point cloud and prototype point cloud from the previous step are voxelized. By comparing the two, if the $i$-th voxel of the instance point cloud contains no points while the corresponding $i$-th voxel of the prototype point cloud contains points, points from the prototype's $i$-th voxel are selected and added to the $i$-th voxel of the instance point cloud. The number of points added is determined by the following formula:
\begin{equation}
num = N_i \times CC,
\end{equation}
where $N_i$ represents the number of points in prototype's $i$-th voxel of the current class.

\begin{figure}[pos=htb]
    \centering
    \begin{minipage}{\linewidth} 
        \centering
        \subfloat[]{\includegraphics[width=0.7\linewidth, trim=0 0 0 0, clip]{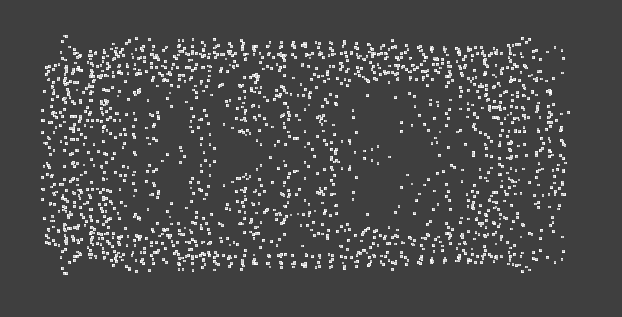}\label{fig_5_case}}
    \end{minipage} \\ 
    \begin{minipage}{\linewidth} 
        \centering
        \subfloat[]{\includegraphics[width=0.7\linewidth, trim=0 0 0 0, clip]{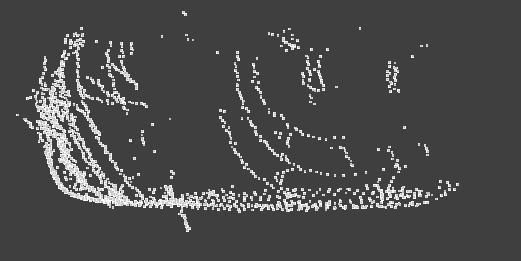}\label{fig_6_case}}
    \end{minipage}\\ 
    \begin{minipage}{\linewidth} 
        \centering
        \subfloat[]{\includegraphics[width=0.7\linewidth, trim=0 0 0 0, clip]{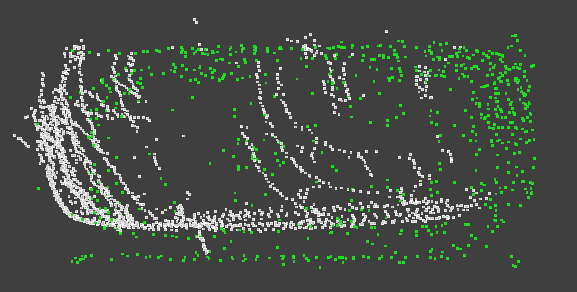}\label{fig_6_case}}
    \end{minipage}
    \caption{(a) shows the prototype point cloud of the car; (b) shows the raw point cloud.(c) illustrates the object's raw point cloud in white and the points generated by the completion module in \textcolor{green}{green}.}
    \label{fig_6}
\end{figure}

A comparison between the prototype point cloud and the completed object point cloud is shown in Fig. \ref{fig_6}.

\begin{figure}[pos=htb]
    \centering
    \subfloat[]{\includegraphics[width=0.3\linewidth, trim=0 0 0 0, clip]{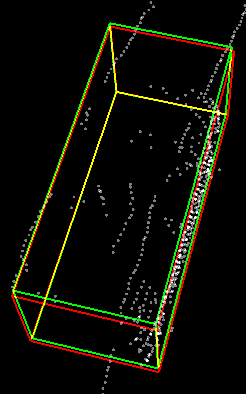}\label{fig_7_case}}
    \hspace{0.05\linewidth} 
    \subfloat[]{\includegraphics[width=0.31\linewidth, trim=0 0 0 0, clip]{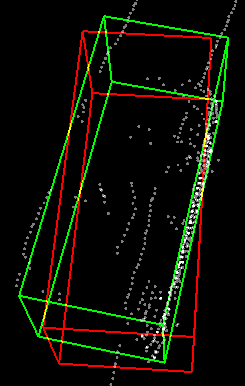}\label{fig_8_case}}
    \caption{In the figure, the \textcolor{green}{red} bounding box represents the detection result, the \textcolor{green}{green} box denotes the ground truth (GT), and the white points depict the raw point cloud. As shown in (a), the detection box generated with the completion module enabled demonstrates closer alignment to the GT box compared to the baseline detection box without the module in (b).}
    \label{fig_7}
\end{figure}

As shown in Fig. \ref{fig_7}, the detection bounding boxes after processing by the completion module align more closely with the ground truth (GT) boxes compared to the baseline without the completion module.

\subsection{Generate Prototype Point Cloud}
The generation of prototype point clouds for different categories in the KITTI dataset must adhere to a key requirement: each category's prototype should comprehensively aggregate the shape characteristics of that category within the dataset while consisting of high-quality points. Our method for generating the prototype point cloud for a specific category proceeds as follows. First, we collect all object point clouds belonging to the target category from the ground truth annotations. These individual point clouds undergo centroid alignment, heading angle rotation, and are translated to the origin, resulting in a consolidated point cloud that encapsulates the complete shape information for the category. To ensure high quality, points with low reflectivity are filtered out from this consolidated cloud. Finally, for all categories, we perform farthest point sampling (FPS) within a defined volumetric range (a horizontal plane from -2 to 2 meters and a vertical height from -1 to 3 meters). Specifically, we sample 2048 points for the vehicle category and 512 points each for bicycle and human categories, yielding the final prototype point clouds. 
\subsection{Loss Function}
The detector with our method is trained using the loss of the baseline model ($L_{bs}$), the loss from the Instance Selection module ($L_I$), and the loss from the Alignment-Based Point Completion module ($L_C$). The total loss is defined as follows:
\begin{equation}
L_{total}= L_I+L_C+L_{bs},
\end{equation}
where $L_I$ represents the Focal Loss \cite{focal} for point classification, which minimizes the classification error of points. The loss $L_C$ for the Alignment-Based Point Completion module is defined as:
\begin{equation}
L_C= \delta L_{center}+L_{angle},
\end{equation}
where $L_{center}$ denotes the $L_1$ loss of centroid offset, used to minimize the offset, and $L_{angle}$ denotes the $L_1$ loss of heading angle, used to minimize the error in the heading angle.
The factor $\delta$ is set to 0.5.
\section{Experiments}
\begin{table*}[width=0.8\textwidth,pos=t]
\caption{ Comparisons on the KITTI validation set.}
\label{KITTI-results}
\scalebox{0.8}{ 
\begin{tabular}{c|c c c|c c c|c c c|c c c}
\hline
\multirow{2}{4em}{Methods} & \multicolumn{3}{c|}{Car 3D AP R40} & \multicolumn{3}{c|}{Ped. 3D AP R40} & \multicolumn{3}{c|}{Cyc. 3D AP R40} & \multicolumn{3}{c}{Average} \\
& Easy & Mod. & Hard & Easy & Mod. & Hard & Easy & Mod. & Hard & Easy & Mod. & Hard \\
\hline
FSDV2 & 90.93 & 81.57 & 76.20 & 59.31 & 53.03 & 47.40 & 83.20 & 62.20 & 58.50 & 77.41 & 65.27 & 61.37 \\
+ SPPCC & \textbf{91.05} & \textbf{81.85} & \textbf{77.00} & \textbf{61.65} & \textbf{55.62} & \textbf{50.36} & \textbf{85.96} & \textbf{64.71} & \textbf{60.56} & \textbf{79.55} & \textbf{67.39} & \textbf{63.64} \\
Improvement & +0.08 & +0.28 & +0.80 & +2.34 & +2.59 & +2.94 & +2.76 & +2.51 & +1.94 & +2.14 & +2.12 & +2.27 \\
\hline
VoxelNext & 85.86 & 76.61 & 75.72 & 54.98 & 51.52 & 48.03 & 83.50 & 65.31 & 63.55 & 74.78 & 64.48 & 62.43 \\
+ SPPCC & \textbf{86.95} & \textbf{77.12} & \textbf{75.72} & \textbf{55.77} & \textbf{52.95} & \textbf{49.66} & \textbf{85.11} & \textbf{67.26} & \textbf{63.74} & \textbf{75.90} & \textbf{65.78} & \textbf{63.05} \\
Improvement & +1.09 & +0.51 & - & +0.79 & +1.43 & +1.63 & +1.61 & +1.95 & +0.19 & +1.12 & +1.30 & +0.62 \\
\hline
\end{tabular}
}
\end{table*}
\begin{table*}[width=0.8\textwidth,pos=t]
\caption{ Comparisons on the KITTI test set.} \label{KITTI-test-results}
\setlength{\belowcaptionskip}{-0.4cm}
\scalebox{0.8}{
\begin{tabular}{c|c c c|c c c|c c c|c c c}
\hline
\multirow{2}{4em}{Methods} & \multicolumn{3}{c|}{Car 3D AP R40} & \multicolumn{3}{c|}{Ped. 3D AP R40} & \multicolumn{3}{c|}{Cyc. 3D AP R40} & \multicolumn{3}{c}{Average} \\
& Easy & Mod. & Hard & Easy & Mod. & Hard & Easy & Mod. & Hard & Easy & Mod. & Hard \\
\hline
FSDV2 & \textbf{86.56} & 75.38 & \textbf{70.98} & 46.33 & 37.70 & 34.50 & 77.53 & 59.09 & 53.39 & 70.14 & 57.39 & 52.95 \\
+ SPPCC & 86.53 & \textbf{75.89} & 70.64 & \textbf{47.21} & \textbf{38.47} & \textbf{35.15} & \textbf{79.14} & \textbf{60.74} & \textbf{55.00} & \textbf{70.96} & \textbf{58.37} & \textbf{53.60} \\
Improvement & - & +0.51 & - & +0.88 & +0.77 & +0.65 & +1.61 & +1.65 & +1.61 & +0.82 & +0.98 & +0.65 \\
\hline
VoxelNext & \textbf{83.88} & 75.58 & 70.77 & 47.46 & 39.97 & 37.43 & 78.18 & 61.74 & 54.68 & 69.84 & 59.09 & 54.29 \\
+ SPPCC & 82.45 & \textbf{75.86} & \textbf{71.26} & \textbf{49.50} & \textbf{41.71} & \textbf{39.14} & \textbf{81.28} & \textbf{62.78} & \textbf{55.78} & \textbf{71.08} & \textbf{60.12} & \textbf{55.39} \\
Improvement & - & +0.28 & +0.49 & +2.04 & +1.74 & +1.71 & +3.10 & +1.04 & +1.10 & +1.24 & +1.03 & +0.86 \\
\hline
\end{tabular}
}

\end{table*}
In this section, we first introduce the datasets and evaluation metrics, followed by a description of the implementation details for the experiments. Next, we present the experimental results obtained on the aforementioned datasets and provide a detailed analysis. Finally, we conduct ablation studies to investigate the contributions of individual components of our proposed method to the overall performance.
\subsection{Datasets and Evaluation Metrics}
Our method is evaluated on the KITTI \cite{KITTI} dataset using both the validation and test sets. The evaluation metric used for KITTI is Average Precision (AP). Detailed information about the dataset and its splits can be found in the supplementary materials.

\subsection{Implementation Details}
We adopted the DBSCAN clustering algorithm as the clusterer in Instance Selection. In the experiments on the KITTI \cite{KITTI} dataset, the ADAM optimizer was used with an initial learning rate of 0.00001. The network was trained on a single NVIDIA 3090 GPU with a batch size of 12 for a total of 60 epochs. When combined with the completion module, the training time for both FSDV2 \cite{FSDV2} and VoxelNext \cite{VoxelNext} was approximately 6 hours. The learning rate scheduling and proposal refinement strategies were consistent with those used in FSDV2. During training, we employed the same 3D object detection data augmentation strategies as the baseline methods.
\subsection{Validation}
As shown in Tab. \ref{KITTI-results} and Tab. \ref{KITTI-test-results}, we report the comparative results on the KITTI \cite{KITTI} validation and test sets. We re-implemented FSDV2 \cite{FSDV2} and VoxelNext \cite{VoxelNext} based on their open-source code and trained our method on top of these two detection frameworks. The results demonstrate that our method improves the performance of both baseline detectors across all categories and difficulty levels.

On the validation set, for FSDV2 , our method achieves an improvement of +2.14\%, +2.12\%, and +1.27\% in the 3D AP at 40 recall positions for the average class under easy, moderate, and hard difficulty levels, respectively. For VoxelNext , our method achieves improvements of +1.12\%, +1.30\%, and +0.62\% in the 3D AP under the same conditions. Moreover, for pedestrian and cyclist categories, the FSDV2-based model achieves an AP improvement of +2.62\% and +2.40\%, respectively, at the average difficulty level, while the VoxelNext-based model achieves improvements of +1.28\% and +1.25\%, respectively. These results collectively indicate that our method significantly enhances the 3D detection performance of the baseline models on the KITTI dataset.

On the test set, for FSDV2 , our method achieves improvements of +0.82\%, +0.98\%, and +0.65\% in the 3D AP at 40 recall positions for the average class under easy, moderate, and hard difficulty levels, respectively. For VoxelNext, under the same conditions, our method improves the 3D AP by +1.24\%, +1.03\%, and +0.86\%, respectively. Additionally, for pedestrian and cyclist categories, the FSDV2-based model achieves an AP improvement of +0.77\% and +1.62\%, respectively, at the average difficulty level, while the VoxelNext-based model achieves improvements of +1.83\% and +1.75\%, respectively. These results validate the effectiveness and generalization ability of our proposed method on the KITTI dataset.

Furthermore, to validate the effectiveness of our method across broader scenarios, we constructed a subset named "waymo-tiny" comprising the first 500 samples from the Waymo validation set as the test set. On this subset, we evaluated both the FSDV2 baseline model trained on the KITTI dataset and the FSDV2 model enhanced with our completion module. As shown in Tab. \ref{waymo-test-results}, the enhanced model demonstrates improved detection performance for vehicles, pedestrians, and cyclists at both Level-1 and Level-2 difficulty metrics.
\begin{table*}[width=0.8\textwidth,pos=t]
\caption{ Comparisons on the waymo-tiny test set.} \label{waymo-test-results}
\setlength{\belowcaptionskip}{-0.4cm}
\begin{tabular}{c|c|c c|c c|c c|c c}
\hline
\multirow{2}{4em}{Methods} &\multirow{2}{4em}{Difficulty}  & \multicolumn{2}{c|}{Vehicle} & \multicolumn{2}{c|}{Pedestrian} & \multicolumn{2}{c|}{Cyclist} & \multicolumn{2}{c}{Average} \\
& & mAP & mAPH & mAP & mAPH & mAP & mAPH & mAP & mAPH \\
\hline
FSDV2 & \multirow{3}{4em}{LEVEL-1}& 68.8 & 68.3 & 73.8 & 68.7 & 69.7& 68.6 & 70.8 & 68.5  \\
+ SPPCC & &  \textbf{69.1} & \textbf{68.9} & \textbf{74.4} & \textbf{69.2} & \textbf{70.0} & \textbf{68.9} & \textbf{71.2} & \textbf{69.0}  \\
Improvement & & +0.3 & +0.6 & +0.6 & +0.5 & +0.3 & +0.3 & +0.4 & +0.5\\
\hline
FSDV2 & \multirow{3}{4em}{LEVEL-2}& 60.4 & 60.0 & 68.4 & 61.5 & 66.9& 65.8 & 65.2 & 62.5  \\
+ SPPCC & &  \textbf{60.7} & \textbf{60.4} & \textbf{68.8} & \textbf{61.7} & \textbf{67.5} & \textbf{66.2} & \textbf{65.7} & \textbf{62.8}  \\
Improvement & & +0.3 & +0.4 & +0.4 & +0.2 & +0.6 & +0.4 & +0.5 & +0.3\\
\hline
\end{tabular}

\end{table*}

\subsection{Ablation Studies}
\begin{table*}[width=0.8\textwidth,pos=t]
\caption{ Ablation studies of Alignment-Based Point Completion.} \label{Ablation studies of Alignment-Based Point Completion.}
\scalebox{0.9}{
\begin{tabular}{c|c|c c|c c c|c }
\hline
\multirow{2}{4em}{Methods} & \multirow{2}{*}{Instance Selection} & \multicolumn{2}{c|}{Alignment-Based Point Completion} & \multicolumn{3}{c|}{Average} & Inference \\
& & Alignment & Completion & Easy & Mod. & Hard & Time (ms) \\
\hline

FSDV2 & - & - & - & 77.41 & 65.27 & 61.37 & \textbf{67} \\
+ SPPCC & \Checkmark & - & \Checkmark & 78.68 & 66.41 & 62.38 & 103 \\
+ SPPCC & \Checkmark & \Checkmark & \Checkmark & \textbf{79.55} & \textbf{67.39} & \textbf{63.64} & 119\\

\hline
\end{tabular}
}
\end{table*}
In the ablation study, we conducted experiments based on FSDV2 to evaluate the design of the Alignment-Based Point Completion Module. The experimental results are presented in Tab. \ref{Ablation studies of Alignment-Based Point Completion.} and \ref{completion coefficient}.

In Tab. \ref{Ablation studies of Alignment-Based Point Completion.}, we performed ablation experiments on the alignment submodule to validate its design rationale. If the alignment step is not used during training and inference in the second stage, meaning the center and yaw angle of the instance point cloud are not aligned with the prototype point cloud and only the centroid of the instance point cloud is aligned with the prototype point cloud, the detection performance decreases, and the inference time increases. When the alignment step is incorporated during training and inference in the second stage, the best AP values (indicated in bold) are achieved across all three difficulty levels. Although the inference time increases compared to the former case, the increase remains within an acceptable range.
\begin{table}[width=0.9\linewidth,pos=h]
\centering
\caption{ Ablation studies of Completion Coefficient ($CC$).} 
\label{completion coefficient}
\begin{tabular}{c|c|c c c }
\hline
\multirow{2}{*}{Methods} & \multirow{2}{*}{$CC$} &\multicolumn{3}{c}{Average} \\
& & Easy & Mod. & Hard  \\
\hline

FSDV2 & -   & 77.41 & 65.27 & 61.37  \\
+ SPPCC & $1$ & 78.63 & 66.22 & 62.00  \\
+ SPPCC & $ 1-N_I /N_p$ & 78.78 & 66.35 & 62.09  \\
+ SPPCC & $N_I /N_p$ & \textbf{79.55} & \textbf{67.39} & \textbf{63.64} \\

\hline
\end{tabular}
\end{table}

In Tab. \ref{completion coefficient}, we conducted ablation experiments on the completion coefficient in the completion submodule to validate its design rationale. We set the completion coefficient to three different values for comparison: when the coefficient is set to $1$, the performance improves slightly, with increases of 1.22\%, 0.95\%, and 0.63\%, respectively. When the coefficient is set to $ 1-N_I /N_p$, the performance improves more significantly, with increases of 1.37\%, 1.08\%, and 0.72\%, respectively. When the coefficient is set to $N_I /N_p$, the best AP values (indicated in bold) are achieved across all three difficulty levels, with relative increases of 2.14\%, 2.12\%, and 2.27\%.

In summary, the design of the Alignment-Based Point Completion Module is reasonable, achieving a good balance between detection accuracy and inference speed.

\section{Conclusion}
In this paper, we propose a point cloud completion method based on Instance Selection and Alignment-Based Point Completion, specifically designed for fully sparse 3D object detection. Specifically, we design a Instance Selection module to identify instances of all categories in the scene, and an Alignment-Based Point Completion module to complete the point cloud. Extensive experimental results demonstrate the effectiveness of our method.


\bibliographystyle{cas-model2-names}

\bibliography{ref}



\end{document}